\documentclass[runningheads]{llncs}
\usepackage{graphicx}
\usepackage{bbm}
\usepackage{floatrow}
\newfloatcommand{capbtabbox}{table}[][\FBwidth]
\usepackage{pgfplots}
\usepackage{wrapfig}
\usepackage{tikz}
\usepackage{comment}
\usepackage{amsmath,amssymb} 
\usepackage{color}
\usepackage{float}
\usepackage[square,numbers]{natbib}

\begin{document}

\pagestyle{headings}
\mainmatter
\def\ECCVSubNumber{7}  

\title{Revisiting the Threat Space for Vision-based Keystroke Inference Attacks} 


\titlerunning{Abbreviated paper title}
%
\author{John Lim \and
True Price \and 
Fabian Monrose  \and 
Jan-Michael Frahm  }
\authorrunning{F. Author et al.}
%
\institute{Department of Computer Science, The University of North Carolina at Chapel Hill, Chapel Hill NC 27514, USA \\
\email{\{jlim13,jtprice,fabian,jmf\}@cs.unc.edu}}
\maketitle

\begin{abstract}
A vision-based keystroke inference attack is a side-channel attack in which an attacker uses an optical device to record users on their mobile devices and infer their keystrokes. The threat space for these attacks has been studied in the past, but we argue that the defining characteristics for this threat space, namely the strength of the attacker, are outdated. Previous works do not study adversaries with vision systems that have been trained with deep neural networks because these models require large amounts of training data and curating such a dataset is expensive. To address this, we create a large-scale synthetic dataset to simulate the attack scenario for a keystroke inference attack. We show that first pre-training on synthetic data, followed by adopting transfer learning techniques on real-life data, increases the performance of our deep learning models. This indicates that these models are able to learn rich, meaningful representations from our synthetic data and that training on the synthetic data can help overcome the issue of having small, real-life datasets for vision-based key stroke inference attacks. For this work, we focus on single keypress classification where the input is a frame of a keypress and the output is a predicted key. We are able to get an accuracy of 95.6\% after pre-training a CNN on our synthetic data and training on a small set of real-life data in an adversarial domain adaptation framework. \textbf{Source Code for Simulator:} https://github.com/jlim13/keystroke-inference-attack-synthetic-dataset-generator-

\end{abstract}

\section{Introduction}
\label{sec:intro}
Mobile devices have become the main interface for many aspects of human life. People use their phones to connect with friends, send work emails, manage personal finances, and capture photos. Not only does the amount of information we channel through our devices increase as our dependence on our devices increases, but so does the level of sensitivity. It is not uncommon for users to enter social security numbers, credit card numbers, birth dates, addresses, or  other private information onto mobile devices. As a result, it remains important to study attacks that threaten mobile privacy and security. It is only by carefully studying the threat landscape can more robust defenses can be devised. 

In this paper, we analyze Vision-based keystroke inference attacks wherein an attacker uses an optical device to record users on their mobile devices and extract user input. In past work, researchers have explored the ability of adversaries to extract information via direct surveillance or reflective surfaces  \cite{backes2009tempest,backes2008compromising,raguram2011ispy,xu2013seeing,balzarotti2008clearshot}, eye gaze \cite{chen2018eyetell}, finger motion \cite{ye2017cracking}, and device perturbations \cite{sun2016visible}. Unfortunately, these works do not examine adversarial settings where the attacker applies deep learning methods | that have revolutionized computer vision in recent years |  and they only consider limited capture scenarios. Consequently, a broad understanding of 
 the threat space for vision-based keystroke inference attacks is missing.
 
\begin{figure}[t!]
    \center{\includegraphics[scale=0.5]
     {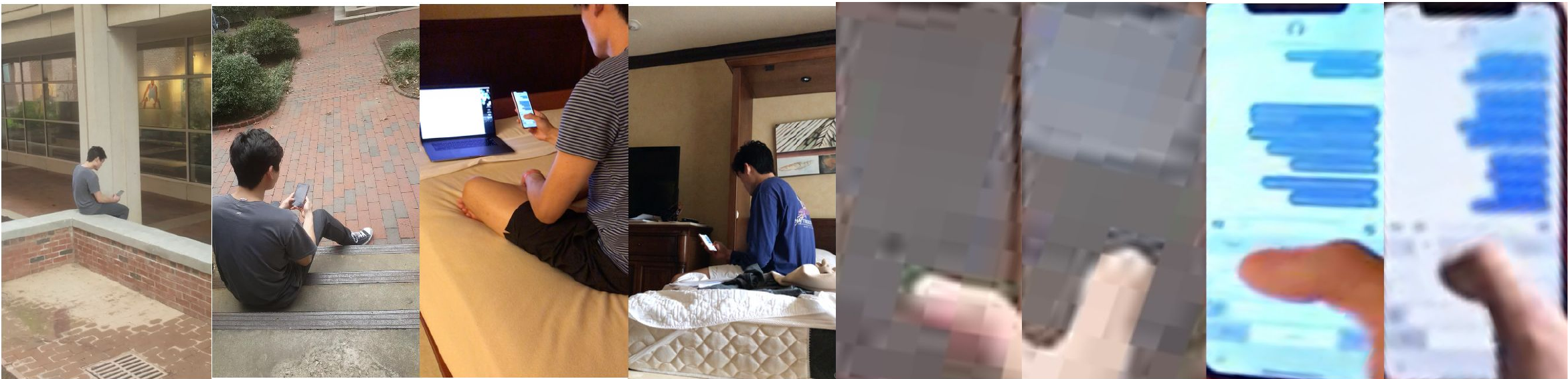} }
    \caption{\label{fig:intro_image} Left: Example of real-life capture scenarios for vision-based keystroke inference attacks. Right: Warped images of the phone to a known template image to consolidate for the various viewpoints.
    }
\end{figure}

To understand the threat posed by deep learning models, we consider the methods by which an attacker might train such a model for general and reliable use.
One of the key factors for the overall success of deep learning is the availability of large, annotated datasets such as ImageNet \cite{deng2009imagenet} and MS COCO \cite{lin2014microsoft}.
Given a large corpus of annotated real-world data (Fig.~\ref{fig:intro_image}), it is reasonable to assume that an attacker could train a powerful model to predict user input from video data.
However, collecting annotated data for vision-based keystroke inference attacks is a \textit{prohibitively} expensive and time-consuming endeavor. Indeed, acquiring a large-enough real-world dataset with sufficient variability for this task would pose a daunting task.
That said, an alternative strategy may be possible: leveraging a simulation engine that offers flexibility to generate a wide array of synthetic, yet realistic, data.

In what follows, we reexamine the threat space for vision-based direct surveillance attacks. Specifically, we examine ability of an adversary equipped with a deep learning systems that feeds off training data created in a systematic way that does not constrain the parameters of the attacker or capture scenario. To do so, we provide a framework for creating a simulation engine that models the capture pipeline of an attacker. Using this framework, we can model different capture scenarios of direct surveillance by permuting the parameters of the simulator: distance, brightness, user's skin tone, angle, capture device, user's device, screen contrast, and typing style. Armed with this framework, we can readily explore the power of adversaries with deep learning capabilities because of the abundance of data we can generate. While there are differences between the synthetic data and real-life data, notably the texture and finger kinematics, we show that our simulator produces data which allows us deep learning models to learn rich, meaningful representations that can be leveraged in the presence of a small set of real-life training data. 

Our specific contributions include:

\begin{itemize}

    \item The first analysis of vision-based keystroke inference attacks with adversaries employing deep learning algorithms and capturing with mobile devices.
    
        
    \item A systematic approach for evaluating inference attacks that can simulate various scenarios by permuting the capture parameters.
 
\end{itemize}

\section{Related Work}
\label{sec:relatedworks}
\subsubsection{Vison-based Keystroke Inference Attacks:}
Some of the earliest works on keystroke inference attacks  focused on direct line of sight and reflective surfaces \textit{(i.e., teapots, sunglasses, eyes)}
 \cite{backes2008compromising,backes2009tempest,raguram2011ispy,xu2013seeing,yue2014blind,ye2017cracking} to infer sensitive data. Under those threat models, an attacker trains a keypress classifier that accounts for various viewing angles and distances by aligning the user's mobile phone to a template keyboard. The success of these attacks rests on the ability to recover graphical pins, words, and full sentences by detecting the individual keypresses. 

More recent work considers threat models where an attacker can not see the screen directly. For example, \citet{sun2016visible} study an attacker who is able to infer keystrokes on an iPad by only observing the back side of the tablet, focusing on the perturbations of the iPad as the user presses a key. They use steerable pyramid decomposition to detect and measure this motion of select areas of interest --- the Apple logo, for example --- in order infer keystrokes. \citet{shukla2014beware}  infer keystrokes by exploiting the spatio-temporal dynamics of a user's hand while typing. No information about the user's screen activity is required, yet it is possible to infer phone, ATM, and door pins. \citet{chen2018eyetell} also create an attack where the user's device is not observed. They track a user's eye gaze to infer graphical pins, iOS pins, and English words. The major drawback of these methods that do not look to exploit the user's on-screen information and finger activity is that attacks do not perform as well compared to the methods that do focus on the user's finger motion and on-screen activity. The adversary trades rate-of-success for discreetness. 

\subsubsection*{Synthetic-to-Real Domain Adaptation:} Synthetic-to-real domain adaptation addresses the dataset bias problem between the synthetic domain $\mathcal{X}^s = \{ \mathbf{x}_i^s, \mathbf{y}_i^s \}$ and real-life domain $\mathcal{X}^t = \{\mathbf{x}_i^t, \mathbf{y}_i^t \}$ where $\mathbf{x}_i \subset \mathbb{R}^d$ is the feature representation and $\mathbf{y}_i$ is the label. $\mathbf{x}_i^s$ and $\mathbf{x}_i^t$ are sampled from two different distributions but share the same feature representation and label space. Computer vision and machine learning algorithms that are trained with supervision require a considerable amount of annotated data that well covers the diverse distribution of application scenarios. Due to the high costs of curating such datasets, many researchers have worked on creating realistic, high-quality synthetic sources. Researchers have developed simulation engines to aid in training algorithms for optical flow \cite{DFIB15}, eye gaze estimation \cite{wood2016learning}, and semantic segmentation \cite{ros2016synthia,Cordts2016Cityscapes,richter2016playing}. Numerous other approaches \cite{bousmalis2017unsupervised,peng2018syn2real,chen2019learning,shrivastava2017learning,hoffman2018cycada} adopt adversarial training to learn a function to produce features that are domain invariant or to transform the pixels in the synthetic data to match distribution of the real data.

\section{Overview}
\label{sec:overview}

\begin{figure*}[htb!]
    \centering
    \includegraphics[width=\textwidth,height=\textheight,keepaspectratio]{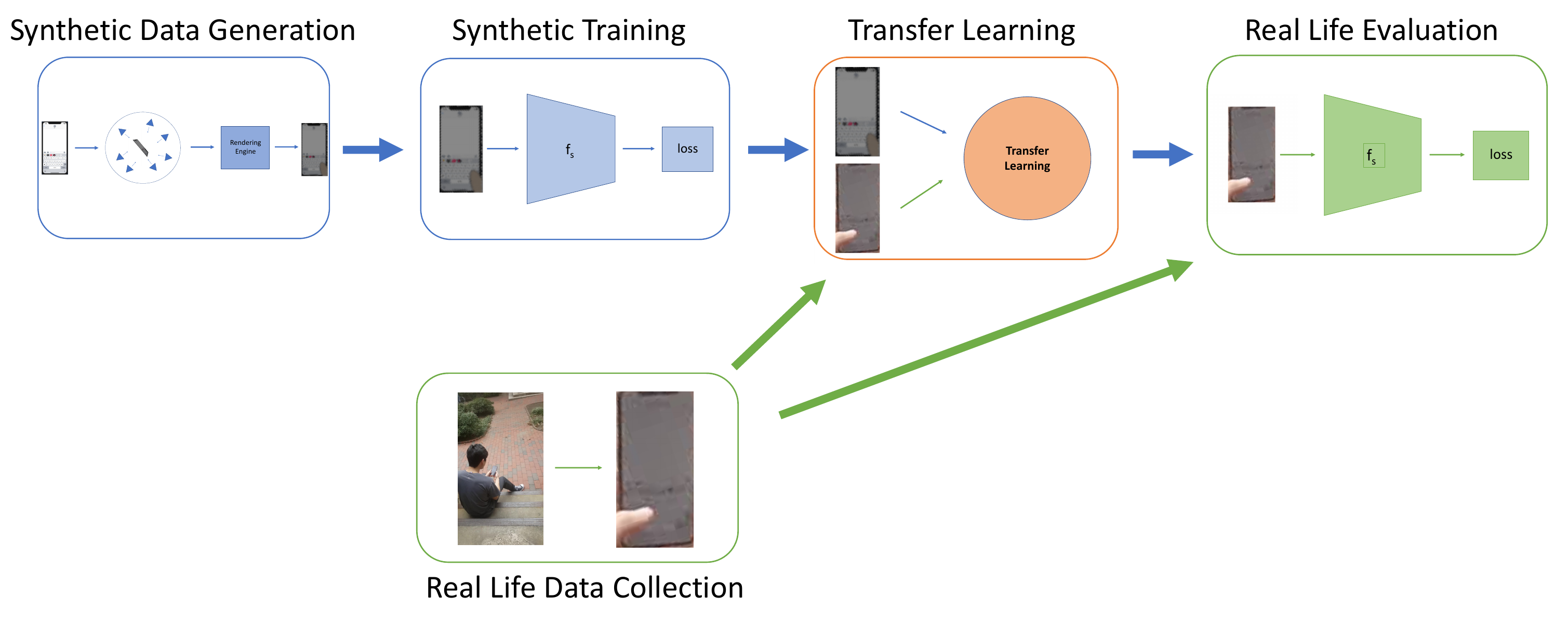}
    \caption{Overview of our Approach. The blue indicates the flow of the synthetic data. The green indicates the flow of the real-life data. The orange indicates where the synthetic-to-real domain transfer learning happens. 
    }
    \label{fig:Overview}
\end{figure*}

The general workflow of our model is highlighted in Fig.~\ref{fig:Overview}. This vision-based keystroke inference attack framework seeks to apply deep learning to a real-world domain in which the attacker has very few labeled datapoints.
First, simulated training data is generated to model the space of attacker parameters, including different viewpoints and recording devices, as well as the victim's texting behavior, for example, finger kinematics.
Note that training annotations come for free with this simulation, as the content of the victim's message is specified by the operator of the simulation engine. During the generation of synthetic data, we also collect and annotate a small set of real-life training data.

After simulation, we train a model, $\mathbf{f_s}$, on just the synthetic data. The representations learned from $\mathbf{f_s}$ are useful in the transfer learning step. $\mathbf{f_s}$ can be fine-tuned with real-life data, if available. Also, $\mathbf{f_s}$ can enforce task consistency when performing pixel-wise domain adaptation techniques \cite{hoffman2018cycada} or the features learned while training the source classifier can be used in adversarial discriminative approaches \cite{tzeng2017adversarial}. Finally, after performing the transfer learning step, the model $\mathbf{f_s}$ can be applied to a real-life test set.

\subsection{Synthetic Dataset Generation}
We develop a simulation engine for keystroke inference attacks in which the attacker has a direct line of sight to the user's phone screen. The parameters that govern our simulations are: the attacker's capture device, user's mobile device, capture distance, screen brightness and contrast, the user's skin tone, and the typed message. Being able to permute these parameters allows us to systematically assess the threat space for this attack because we are not restricted to a fixed attack setting. The general pipeline is displayed in the ``Synthetic Data Generation'' module of Fig.~\ref{fig:Overview}. 

\subsubsection{Capture Stage}
In the capture stage, the attacker uses an optical device such as a mobile phone's camera to record the user's behavior. In our scenario, the attacker focuses on recording the user's device and the associated finger movements that result from different keypresses. For our experiments, we set the attacker's capture device as the camera on the iPhone 6 and the user's device as an iPhone XR. For simulation, we utilize 3D models of the iPhone XR and the user's thumb.
For a given keypress, we align the thumb model over the associated key and then render the thumb and iPhone models into a randomly selected attacker viewpoint, thereby simulating what a real-world attacker would observe.

\subsubsection{Information Extraction and Alignment}
This critical step allows us to consolidate all of the varying capturing positions of the attacker to one view. Given an image of the user's finger and phone (from the attacker's point of view), we need to extract  meaningful information such as the type of phone or localization of the finger. This can be done via computer vision algorithms, for example, running a phone detection algorithm localize the phone, or by manually cropping out the phone in the image. Regardless of the approach, the goal of this step is to extract the most salient information from the given image. In our case, we assume that the attacker can manually crops out the phone the most salient information is the phone and user's fingers. 

Next, we need to extract the four corners of the user's device in order to align the image of the phone to a reference template via a homography. A homography is a a 2D projective transformation that relates two images of the same planar object. The phone is a planar object that is captured from varying viewpoints. The phone in the images that we capture are all related to each other by a homography transformation. Given any image of the phone, we can warp that image to a template image by the homography matrix, $\mathbf{H}$. We can calculate $\mathbf{H}$ using the four corners of the rendered image, which we know from simulation, and the four corners of the template image. In our explorations, we use the iPhone XR image from Apple Xcode's simulator as our template image and use the dimensions of the phone, which are available online, for the four corners. While the captured phone and template phone are both planar objects, the thumb is not. There is minimal distortion and our experiments show that this does not affect the learning process. Once the captured image and template image are aligned, we can train a classifier to predict the keypress. The input to the classifier is a single image of a single key press, and the output is a prediction of which key was pressed.

\begin{figure}
    \center{\includegraphics[scale=0.5]
     {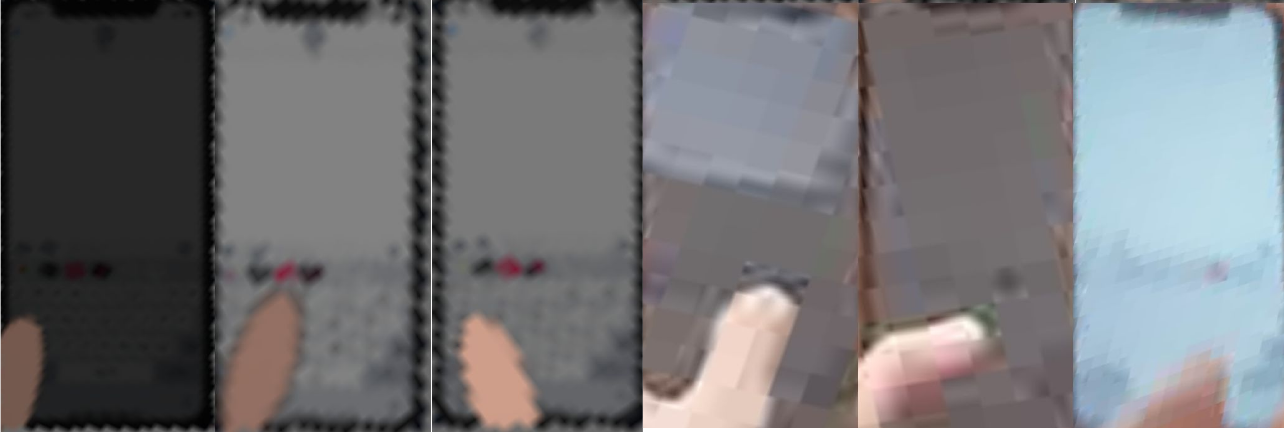} }
    \caption{\label{fig:sample_images} Left: Examples of our synthetic data. Right: Examples of our real-life data. 
    }
\end{figure}

\subsection{Single Keypress Classification}

For single keypress classification, the input is a single image of the user's thumb over a specific key. As previously mentioned, the input image is homography-aligned to the reference image to consolidate the different viewpoints from which the attacker can capture the user. We train a model to output a predicted key, $p^{'}$ on a QWERTY keyboard.

We train a Logistic Regression, Support Vector Machine (SVM) and a Convolutional Neural Network (CNN) for our evaluations, and show that the deep learning approach performs better than the shallow methods. This is a $26$-way classification task that is trained by the cross entropy loss function:

\begin{equation}
  \label{cls_equation}
  L = E_{(x,y)\sim X}\sum_{n=1}^{N} \mathbbm{1}_{[n = y]} \log(\sigma(f(x)))
\end{equation}

Previous works have analyzed single keypress classification, but we differ in that we only focus on attackers with mobile devices. \citet{xu2013seeing} and \citet{raguram2011ispy} use high-end recording devices in their setup, and while those devices are considerably smaller and cheaper than telescopes used by \citet{backes2008compromising,backes2009tempest} and \citet{kuhn2002compromising}, the size and conspicuousness of such devices still restricts their use in discreet capturing scenarios. While mobile cameras have less capture capability than a high-end DSLR or telescope, they allow for more discreet capturing, making the attack less noticeable.

\subsubsection{Transfer Learning}

We adopt transfer learning techniques to bridge the gap between the synthetic and real-life data distributions. Recall that the majority of our data comes from a simulation engine and we do the majority of our training on this data to compensate for the difficulty in collecting real-life data samples. We adopt two approaches for transfer learning: fine-tuning and adversarial domain adaptation, similar to the technique introduced by \citet{tzeng2017adversarial}.

CNNs are successful in vision tasks in which we have large amounts of training data because they are able to learn powerful representations with millions of parameters.
Due to the high number of parameters in these CNNs, we are unable to learn meaningful representations on a small dataset. \citet{oquab2014learning} and \citet{chu2016best}
have shown that CNNs initially trained on a large dataset can transfer those representations to the target dataset by fine-tuning the CNN. Fine-tuning allows us to learn the key features for the general task of single key press classification using synthetic data and allows adjust the weights for a specific domain of single key press classification. Our results show that a CNN trained on our synthetic data learns a representation that can be transferred to real life data.

We follow the Adversarial Discriminative Domain Adaptation, ADDA, framework introduced by \citet{tzeng2017adversarial} where the purpose is to learn a domain invariant feature representation between the source and target domain. The source domain is denoted as $\mathcal{X}^s = \{ \mathbf{x}_i^s, \mathbf{y}_i^s \}$ where $\mathbf{x}_i^s$ is the feature representation and $\mathbf{y}_i^s$ is the label. The target domain is denoted as $\mathcal{X}^t = \{ \mathbf{x}_i^t, \mathbf{y}_i^t \}$. In the target domain we have a small set of labeled instances. A visual classifier, $f$, can be decomposed into two functions, $f = g \circ h$. $g$ is the feature extractor that takes the input image into a d-dimensional feature space and $h$ is the predictor that takes the feature representation and outputs a probability distribution over the label space. $g_s$ and $g_t$ represent the feature extractors for the source and target domains, respectively. $h_s$ and $h_t$ represent the predictors.

The training for ADDA is done in multiple stages; we do make some slight adjustments to the training process as we have access to a small set of labels in the target domain. First, we train $g_s$ and $h_s$ to minimize the loss function \ref{cls_equation}. Next, we train $g_t$ in an adversarial fashion. We maximize the discriminator's ability to distinguish between the features outputted from $g_t$ and $g_s$ while also forcing $g_t$ to extract features that are indistinguishable from those extracted from $g_s$. During this step, we also train $h_t$. Finally, we test using $g_t$ and $h_t$. The optimization procedure is formally denoted below: 

\begin{equation}
  \label{cls_equation_source}
  \min_{g_s, h_s} \mathcal{L}_{cls} = E_{(x_s,y_s)\sim X_s}\sum_{n=1}^{N} \boldsymbol{1}_{[n = y_s]} \log(\mathit{\sigma}( h_s(g_s(x_s))))
\end{equation}

\begin{equation}
    \label{gan_eq}
      \min_D \mathcal{L}_{adv}(X_s, X_t, g_s, g_t) = E_{x_s\sim X_s} \, [\log \, g_s(x_s)] +  E_{x_t \sim X_t } \, [\log \, (1 − g_t(x_t)) ] 
\end{equation}
\begin{equation}
    \label{feat_ext}
      \min_{g_s, g_t} \mathcal{L}_{g}(X_s, X_t, D) = E_{x_t\sim X_t} \, [\log \, D(g_t(x_t))] 
\end{equation}

\begin{equation}
  \label{cls_equation_target}
  \min_{g_t, h_t} \mathcal{L}_{cls} = E_{(x_t,y_t)\sim X_t}\sum_{n=1}^{N} 1_{[n = y_t]} \log(\mathit{\sigma}( h_t(g_t(x_t))))
\end{equation}

\ref{cls_equation_source} is trained by itself in the first stage. \ref{gan_eq} and \ref{feat_ext} are trained together in the next step, and when labels are available, \ref{cls_equation_target} is also used. 

\section{Experiments}
\label{sec:experiments}

Next, we share our implementation details and experimental setup for single keypress classification. We also utilize our simulation engine to simulate different types of defenses and evaluate their effectiveness.

\subsection{Single Keypress Classification}
Perhaps not surprisingly, we show that a deep learning based approach outperforms shallow machine learning methods for single keypress classification. The input to these models is a single frame of a keypress and the output is a prediction of the pressed key. Our synthetic data consists of 15,000 keypress images and were labeled as one of the 26 letters in the alphabet. We randomized the lighting, screen blur, screen contrast, camera angle, distance (1-7 meters), and skin tone in order to simulate various capture conditions and to diversify our dataset. Our real life training data consisted of 540 images of single key press frames. The dataset is split to 390, 80, and 80 images for training, testing, and validation, respectively. These images were captured at distances of up to 5 meters and were taken in both indoor and outdoor settings. We show that our accuracy on our real-life test set is similar to that of our synthetic data after adopting transfer learning techniques. We also conduct experiments to see how the minimum number of instances for each class affects transfer learning. The full real-life dataset has 15 instances for each class. 

\begin{table}
\begin{tabular}{ |c|c|c|c|} 
\hline
Method & Synthetic & Real-Life   \\
\hline
Logistic Regression & 81.8\% & 78.3\% \\ 
SVM & 80.4\%  & 75.1\% \\ 
CNN  & 96.3\%  &  76.2\%  \\ 
\hline

\end{tabular}
\caption{\label{Single Key Class}Single Key Classification. The scores under the Synthetic column are trained and evaluated on only synthetic data. Similarly, for the Real-Life column. }
\end{table}
\begin{table}
\begin{tabular}{ |c|c|c|c|c|} 
\hline
 & Real Only & Finetuning & ADDA   \\
\hline
CNN  & 76.2\%  & 93.08\%  &  95.6\%  \\ 
\hline
\end{tabular}
\caption{\label{single key real life data}Single Key Classification on real-life data. We compare the performance of a CNN on a real-life test set. No Adaptation means that the CNN has a random weight initialization and is trained using the real-life training set only. Finetuning and ADDA use the CNN trained on synthetic as the initialization.  }
\end{table}

\subsubsection{Experimental Setup}
We use a linear regression and a SVM as our baseline methods for this task. A 3-layer CNN is used for our deep learning model. Each layer follows a Conv2d-BatchNorm2d-ReLU-MaxPool structure. Each layer has filters of size $5 x 5$, stride 1, and padding 2. The channels are 16, 32, and 16 for each layer. After these convolution layers, there is a linear layer, followed by a 26-way softmax layer. We set our initial learning rate to 0.0002 and use the Adam optimizer. We also crop out the image so that we are only focusing on the keyboard and location of the finger.

For the finetuning experiments, we take the same CNN architecture trained just on synthetic data, replace the last linear layer with a new one, and freeze the early layers. We used a learning rate of 0.00002 using the Adam optimizer and trained it for 60 epochs. For the ADDA results, we use a learning rates of 0.0002 and 0.0004 for the classifiers and feature extractors, respectively.

\subsubsection{Results}
Table \ref{Single Key Class} shows that the CNN significantly outperforms the two shallow methods when trained and evaluated synthetic data. However, if we train and evaluate on just the real-life data, we see a decrease in performance because the model is overfitting to the training data. Training and evaulating on such a small dataset does not give us any insight into this attack because a dedicated attacker could curate his own dataset large enough to benefit from deep learning approaches. We adopt transfer learning approaches to compensate for our lack of real-life training data. Finetuning gives us a classification score that approaches the synthetic data performance, which indicates that our simulation engine is capable of generating data to evaluate single key press classification when we are constrained with limited real-life data. ADDA yields the highest results. In \ref{plot:fewshot}, we show our performance on the real-life test while decreasing the number of per-class examples seen during training.



\begin{center}
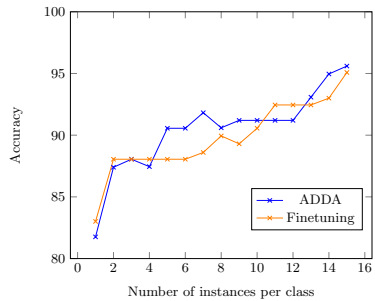
\begin{figure}
\resizebox{5cm}{4cm}{%
    \begin{tikzpicture}
      \begin{axis}[
          ymin=80,
          ymax=100,
          xlabel=Number of instances per class,
          ylabel=Accuracy,
          legend style={at={(0.6,0.2)},
          anchor=west,legend columns=1},
          legend style={font=\small},
          legend entries={ADDA,
                          Finetuning
                          }
         ]

        \addplot[mark=x,blue] coordinates {
          (1, 81.76)
          (2, 87.4)
          (3, 88.05)
          (4, 87.45)
          (5, 90.56)
          (6, 90.56)
          (7, 91.82)
          (8, 90.6)
          (9, 91.2)
          (10, 91.2)
          (11, 91.2)
          (12, 91.2)
          (13, 93.08)
          (14, 94.96)
          (15, 95.6)
        };
         \addplot[mark=x,orange] coordinates {
          (1, 83.01)
          (2, 88.05)
          (3, 88.05)
          (4, 88.05)
          (5, 88.05)
          (6, 88.05)
          (7, 88.6)
          (8, 89.94)
          (9, 89.3)
          (10, 90.56)
          (11, 92.45)
          (12, 92.45)
          (13, 92.45)
          (14, 93)
          (15, 95.08)
        };
        
    \end{axis}
    \end{tikzpicture}
}
\caption{The accuracy on the real-life test set is plotted against the number of per-class instances seen during training.}
\end{figure}
\label{plot:fewshot}
\end{center}

\subsection{Defenses}

Establishing defenses that generalize against multiple attacks is a very challenging problem. One of the main challenges for establishing defenses for vision-based keystroke inference attacks is that there are a few number of methods to prevent an attacker from capturing a user's behavior. The most effective method is abstaining from mobile phone usage, but that is not a practical solution for most people. Many of the previous suggest countermeasures to the attacks it presented, but a defense for attack A can be the threat scenario for attack B. For example, the defense against an attacker who exploits eye gaze would be to wear dark, protective eyewear such as sunglasses, but the user is at risk against an attacker who looks for compromising reflections. Some defenses that can generalize to multiple attacks is the user of randomized keyboards, typing fast, and dynamically moving the phone while typing. While we are not able to study all possible defenses, our simulation engine allows to study some subset of defenses in a systematic way. 

\begin{figure}
\begin{floatrow}
\ffigbox{%
}{%
 {\includegraphics[scale=0.24]
         {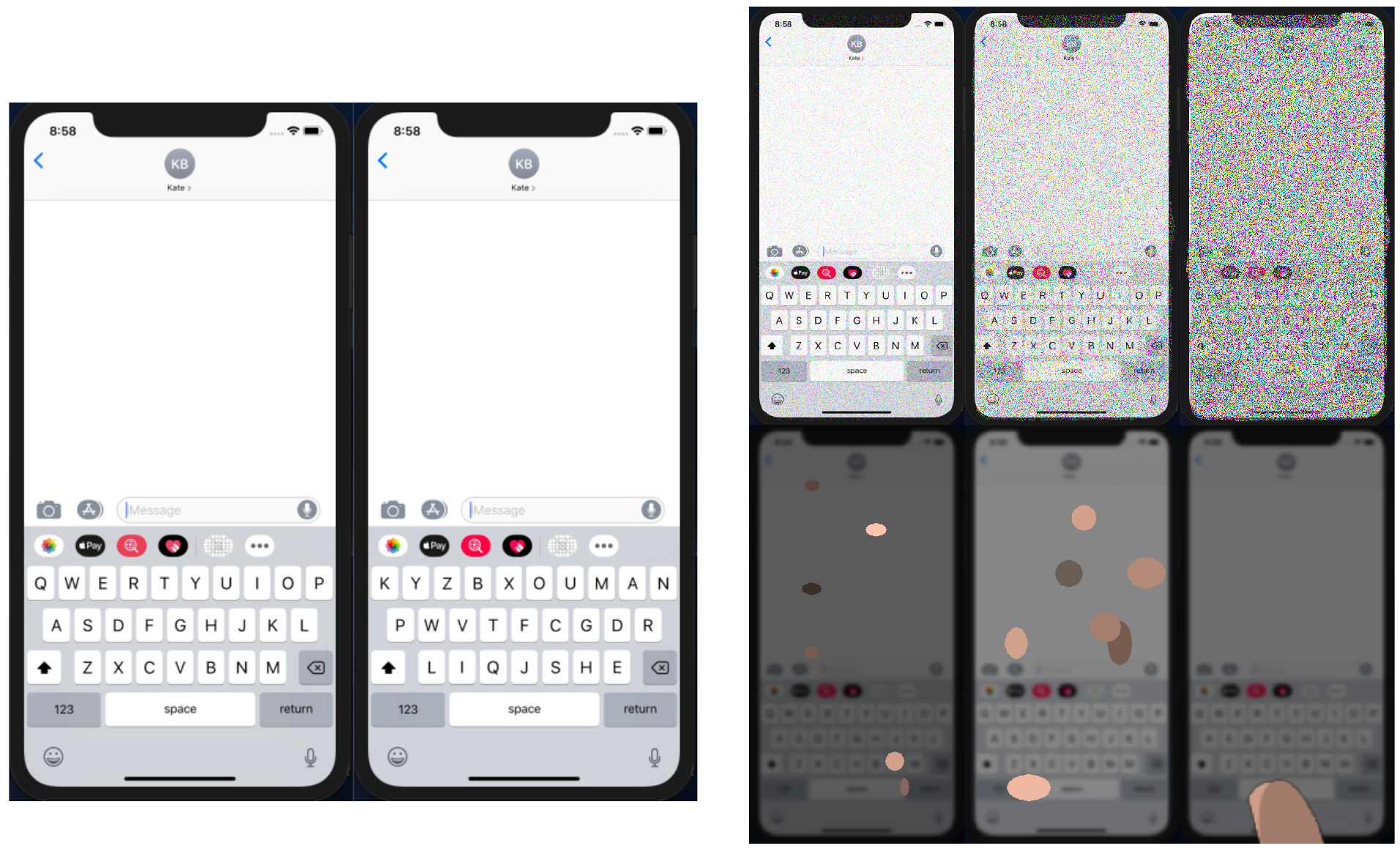} }
        \caption{\label{fig:defenses} Left: the standard QWERTY keyboard. Next to it is a randomly permuted keyboard. For our experiments, note that we only permute the 26 letters. Right: different on-screen perturbations. The top phone screens are corrupted with Gaussian noise with $\sigma$s  of 0.01, 0.05, and 0.75, respectively. The bottoms have on-screen blocks of varying colors and sizes}
}
\capbtabbox{%
  \begin{tabular}{ |c|c|c|} 
\hline
On-Screen Perturbation & Accuracy \\
\hline
Gaussian ($\sigma$ = 0.001) & 94.3\% \\
Gaussian ($\sigma$ = 0.05) & 94.3\%  \\
Gaussian ($\sigma$ = 0.15) & 94.3\%  \\
Gaussian ($\sigma$ = 0.75) & 92\% \\
Small Corruption & 82.1\%  \\
Large Corruption & 45.4\%  \\
Thumb Corruption & 84.8\%  \\
\hline
\end{tabular}
}{\caption{\label{perturbations}Evaluating On-Screen Perturbations as Defenses. We evaluate how a CNN trained without ever seeing any of these perturbations performs against them.}%
}
\end{floatrow}
\end{figure}

\subsubsection{Experimental Setup}

We simulate a few possible on-screen perturbations as defenses for single key press classification. These perturbations can be emitted from the mobile phone's screen. We simulate different Gaussian noise patterns and different "phone screen corruptions" as types of on-screen perturbations. Small and Large Corruptions are those in which we have the phone randomly emit various shapes across the phone screen. A Thumb Corruption is when the phone emits a random shape around the user's thumb when he presses a key. We also study how our methods perform against randomized keyboards. 

Randomized keyboards are one of the most effective ways to defend against keystroke inference attacks. Vision-based keystroke inference attacks learn a mapping between user behavior to a fixed keyboard layout. This mapping is broken when the keyboard layout changes. We generate a dataset of randomized keyboards and evaluate how a CNN trained on layout A performs on layout B.

\subsubsection{Results}

To evaluate the different on-screen perturbations, we first train a CNN on synthetic data without any of these perturbations. Then, we evaluate the model on a separate test set with these corruptions. The results are displayed in \ref{fig:defenses}. The defense becomes more effective as the severity of the perturbations increase. Of course, doing so takes away from the user's usability. While some of the on-screen perturbations were effective in spoofing the CNN, if we were to train the CNN with these perturbations in the training set, the defenses do not hold. The CNN is unaffected as these perturbations become a form of data augmentation. 

Randomized keyboards are an effective defense against the model used in our experiments. We first train a CNN on synthetic data on a QWERTY layout. Then, we generate a training set with a fixed layout, B, that is \textbf{not} a QWERTY. If we evaluate the CNN trained on the QWERTY, we do not get better than $0.04 \%$ accuracy, which means that the classifier is guessing. Similarly, if we generate a new training and testing set, all with instances of randomly permuted keyboards, we still do not do better than guessing. This indicates that the model can not recover any sort of information from the keyboard to indicate which key is pressed. The model effectively learns a mapping from the user's finger tip to an assumed keyboard layout. If that assumption is broken, then the model can not predict the key. Of course, the biggest sacrifice for using a randomized keyboard is the severe decrease in usability.

\section{Conclusion}
We explored a method to evaluate deep learning methods for vision based keystroke inference attacks; a domain in which curating a dataset large and diverse enough for deep learning methods is expensive. 
In doing so, we developed a simulation engine that generates data that allows us to systematically study these attacks by manipulating various parameters (e.g., capture distance, camera rotation, screen brightness, texting speed). Similarly, this capability allows us to study different defenses. 
We create synthetic data for the task of single key press classification, and show that deep learning models, when pre-trained on this data, are able to learn powerful representations that compensate for the lack of real-life training data. 
Our experiments not only show that deep learning approach outperforms shallow methods for single key press classification, but also show that an attacker does not need many real-life data points to train such a classifier. These experiments indicate that we need to rethink our beliefs of the threat space for vision-based keystroke inference attacks, as they are outdated.

\clearpage

\bibliographystyle{plainnat}

\bibliography{egbib}

\begin{thebibliography}{26}
\providecommand{\natexlab}[1]{#1}
\providecommand{\url}[1]{\texttt{#1}}
\expandafter\ifx\csname urlstyle\endcsname\relax
  \providecommand{\doi}[1]{doi: #1}\else
  \providecommand{\doi}{doi: \begingroup \urlstyle{rm}\Url}\fi

\bibitem[Backes et~al.(2008)Backes, D{\"u}rmuth, and
  Unruh]{backes2008compromising}
Michael Backes, Markus D{\"u}rmuth, and Dominique Unruh.
\newblock Compromising reflections-or-how to read lcd monitors around the
  corner.
\newblock In \emph{2008 IEEE Symposium on Security and Privacy (sp 2008)},
  pages 158--169. IEEE, 2008.

\bibitem[Backes et~al.(2009)Backes, Chen, Duermuth, Lensch, and
  Welk]{backes2009tempest}
Michael Backes, Tongbo Chen, Markus Duermuth, Hendrik~PA Lensch, and Martin
  Welk.
\newblock Tempest in a teapot: Compromising reflections revisited.
\newblock In \emph{2009 30th IEEE Symposium on Security and Privacy}, pages
  315--327. IEEE, 2009.

\bibitem[Balzarotti et~al.(2008)Balzarotti, Cova, and
  Vigna]{balzarotti2008clearshot}
Davide Balzarotti, Marco Cova, and Giovanni Vigna.
\newblock Clearshot: Eavesdropping on keyboard input from video.
\newblock In \emph{2008 IEEE Symposium on Security and Privacy (sp 2008)},
  pages 170--183. IEEE, 2008.

\bibitem[Bousmalis et~al.(2017)Bousmalis, Silberman, Dohan, Erhan, and
  Krishnan]{bousmalis2017unsupervised}
Konstantinos Bousmalis, Nathan Silberman, David Dohan, Dumitru Erhan, and Dilip
  Krishnan.
\newblock Unsupervised pixel-level domain adaptation with generative
  adversarial networks.
\newblock In \emph{Proceedings of the IEEE conference on computer vision and
  pattern recognition}, pages 3722--3731, 2017.

\bibitem[Chen et~al.(2018)Chen, Li, Zhang, Zhang, and
  Hedgpeth]{chen2018eyetell}
Yimin Chen, Tao Li, Rui Zhang, Yanchao Zhang, and Terri Hedgpeth.
\newblock Eyetell: Video-assisted touchscreen keystroke inference from eye
  movements.
\newblock In \emph{2018 IEEE Symposium on Security and Privacy (SP)}, pages
  144--160. IEEE, 2018.

\bibitem[Chen et~al.(2019)Chen, Li, Chen, and Gool]{chen2019learning}
Yuhua Chen, Wen Li, Xiaoran Chen, and Luc~Van Gool.
\newblock Learning semantic segmentation from synthetic data: A geometrically
  guided input-output adaptation approach.
\newblock In \emph{Proceedings of the IEEE Conference on Computer Vision and
  Pattern Recognition}, pages 1841--1850, 2019.

\bibitem[Chu et~al.(2016)Chu, Madhavan, Beijbom, Hoffman, and
  Darrell]{chu2016best}
Brian Chu, Vashisht Madhavan, Oscar Beijbom, Judy Hoffman, and Trevor Darrell.
\newblock Best practices for fine-tuning visual classifiers to new domains.
\newblock In \emph{European conference on computer vision}, pages 435--442.
  Springer, 2016.

\bibitem[Cordts et~al.(2016)Cordts, Omran, Ramos, Rehfeld, Enzweiler, Benenson,
  Franke, Roth, and Schiele]{Cordts2016Cityscapes}
Marius Cordts, Mohamed Omran, Sebastian Ramos, Timo Rehfeld, Markus Enzweiler,
  Rodrigo Benenson, Uwe Franke, Stefan Roth, and Bernt Schiele.
\newblock The cityscapes dataset for semantic urban scene understanding.
\newblock In \emph{Proc. of the IEEE Conference on Computer Vision and Pattern
  Recognition (CVPR)}, 2016.

\bibitem[Deng et~al.(2009)Deng, Dong, Socher, Li, Li, and
  Fei-Fei]{deng2009imagenet}
Jia Deng, Wei Dong, Richard Socher, Li-Jia Li, Kai Li, and Li~Fei-Fei.
\newblock Imagenet: A large-scale hierarchical image database.
\newblock 2009.

\bibitem[Dosovitskiy et~al.(2015)Dosovitskiy, Fischer, Ilg, H{\"a}usser,
  Haz{\i}rba{\c{s}}, Golkov, v.d. Smagt, Cremers, and Brox]{DFIB15}
A.~Dosovitskiy, P.~Fischer, E.~Ilg, P.~H{\"a}usser, C.~Haz{\i}rba{\c{s}},
  V.~Golkov, P.~v.d. Smagt, D.~Cremers, and T.~Brox.
\newblock Flownet: Learning optical flow with convolutional networks.
\newblock In \emph{IEEE International Conference on Computer Vision (ICCV)},
  2015.
\newblock URL
  \url{http://lmb.informatik.uni-freiburg.de/Publications/2015/DFIB15}.

\bibitem[Hoffman et~al.(2018)Hoffman, Tzeng, Park, Zhu, Isola, Saenko, Efros,
  and Darrell]{hoffman2018cycada}
Judy Hoffman, Eric Tzeng, Taesung Park, Jun-Yan Zhu, Phillip Isola, Kate
  Saenko, Alyosha Efros, and Trevor Darrell.
\newblock Cy{CADA}: Cycle-consistent adversarial domain adaptation, 2018.
\newblock URL \url{https://openreview.net/forum?id=SktLlGbRZ}.

\bibitem[Kuhn(2002)]{kuhn2002compromising}
Markus~Guenther Kuhn.
\newblock \emph{Compromising emanations: eavesdropping risks of computer
  displays}.
\newblock PhD thesis, University of Cambridge, 2002.

\bibitem[Lin et~al.(2014)Lin, Maire, Belongie, Hays, Perona, Ramanan,
  Doll{\'a}r, and Zitnick]{lin2014microsoft}
Tsung-Yi Lin, Michael Maire, Serge Belongie, James Hays, Pietro Perona, Deva
  Ramanan, Piotr Doll{\'a}r, and C~Lawrence Zitnick.
\newblock Microsoft coco: Common objects in context.
\newblock In \emph{European conference on computer vision}, pages 740--755.
  Springer, 2014.

\bibitem[Oquab et~al.(2014)Oquab, Bottou, Laptev, and Sivic]{oquab2014learning}
Maxime Oquab, Leon Bottou, Ivan Laptev, and Josef Sivic.
\newblock Learning and transferring mid-level image representations using
  convolutional neural networks.
\newblock In \emph{Proceedings of the IEEE conference on computer vision and
  pattern recognition}, pages 1717--1724, 2014.

\bibitem[Peng et~al.(2018)Peng, Usman, Saito, Kaushik, Hoffman, and
  Saenko]{peng2018syn2real}
Xingchao Peng, Ben Usman, Kuniaki Saito, Neela Kaushik, Judy Hoffman, and Kate
  Saenko.
\newblock Syn2real: A new benchmark forsynthetic-to-real visual domain
  adaptation.
\newblock \emph{arXiv preprint arXiv:1806.09755}, 2018.

\bibitem[Raguram et~al.(2011)Raguram, White, Goswami, Monrose, and
  Frahm]{raguram2011ispy}
Rahul Raguram, Andrew~M White, Dibyendusekhar Goswami, Fabian Monrose, and
  Jan-Michael Frahm.
\newblock ispy: automatic reconstruction of typed input from compromising
  reflections.
\newblock In \emph{Proceedings of the 18th ACM conference on Computer and
  communications security}, pages 527--536. ACM, 2011.

\bibitem[Richter et~al.(2016)Richter, Vineet, Roth, and
  Koltun]{richter2016playing}
Stephan~R Richter, Vibhav Vineet, Stefan Roth, and Vladlen Koltun.
\newblock Playing for data: Ground truth from computer games.
\newblock In \emph{European conference on computer vision}, pages 102--118.
  Springer, 2016.

\bibitem[Ros et~al.(2016)Ros, Sellart, Materzynska, Vazquez, and
  Lopez]{ros2016synthia}
German Ros, Laura Sellart, Joanna Materzynska, David Vazquez, and Antonio~M
  Lopez.
\newblock The synthia dataset: A large collection of synthetic images for
  semantic segmentation of urban scenes.
\newblock In \emph{Proceedings of the IEEE conference on computer vision and
  pattern recognition}, pages 3234--3243, 2016.

\bibitem[Shrivastava et~al.(2017)Shrivastava, Pfister, Tuzel, Susskind, Wang,
  and Webb]{shrivastava2017learning}
Ashish Shrivastava, Tomas Pfister, Oncel Tuzel, Joshua Susskind, Wenda Wang,
  and Russell Webb.
\newblock Learning from simulated and unsupervised images through adversarial
  training.
\newblock In \emph{Proceedings of the IEEE Conference on Computer Vision and
  Pattern Recognition}, pages 2107--2116, 2017.

\bibitem[Shukla et~al.(2014)Shukla, Kumar, Serwadda, and
  Phoha]{shukla2014beware}
Diksha Shukla, Rajesh Kumar, Abdul Serwadda, and Vir~V Phoha.
\newblock Beware, your hands reveal your secrets!
\newblock In \emph{Proceedings of the 2014 ACM SIGSAC Conference on Computer
  and Communications Security}, pages 904--917. ACM, 2014.

\bibitem[Sun et~al.(2016)Sun, Jin, Chen, Zhang, Zhang, and
  Zhang]{sun2016visible}
Jingchao Sun, Xiaocong Jin, Yimin Chen, Jinxue Zhang, Yanchao Zhang, and Rui
  Zhang.
\newblock Visible: Video-assisted keystroke inference from tablet backside
  motion.
\newblock In \emph{NDSS}, 2016.

\bibitem[Tzeng et~al.(2017)Tzeng, Hoffman, Saenko, and
  Darrell]{tzeng2017adversarial}
Eric Tzeng, Judy Hoffman, Kate Saenko, and Trevor Darrell.
\newblock Adversarial discriminative domain adaptation.
\newblock In \emph{Proceedings of the IEEE Conference on Computer Vision and
  Pattern Recognition}, pages 7167--7176, 2017.

\bibitem[Wood et~al.(2016)Wood, Baltru{\v{s}}aitis, Morency, Robinson, and
  Bulling]{wood2016learning}
Erroll Wood, Tadas Baltru{\v{s}}aitis, Louis-Philippe Morency, Peter Robinson,
  and Andreas Bulling.
\newblock Learning an appearance-based gaze estimator from one million
  synthesised images.
\newblock In \emph{Proceedings of the Ninth Biennial ACM Symposium on Eye
  Tracking Research \& Applications}, pages 131--138. ACM, 2016.

\bibitem[Xu et~al.(2013)Xu, Heinly, White, Monrose, and Frahm]{xu2013seeing}
Yi~Xu, Jared Heinly, Andrew~M White, Fabian Monrose, and Jan-Michael Frahm.
\newblock Seeing double: Reconstructing obscured typed input from repeated
  compromising reflections.
\newblock In \emph{Proceedings of the 2013 ACM SIGSAC conference on Computer \&
  communications security}, pages 1063--1074. ACM, 2013.

\bibitem[Ye et~al.(2017)Ye, Tang, Fang, Chen, Kim, Taylor, and
  Wang]{ye2017cracking}
Guixin Ye, Zhanyong Tang, Dingyi Fang, Xiaojiang Chen, Kwang~In Kim, Ben
  Taylor, and Zheng Wang.
\newblock Cracking android pattern lock in five attempts.
\newblock 2017.

\bibitem[Yue et~al.(2014)Yue, Ling, Fu, Liu, Ren, and Zhao]{yue2014blind}
Qinggang Yue, Zhen Ling, Xinwen Fu, Benyuan Liu, Kui Ren, and Wei Zhao.
\newblock Blind recognition of touched keys on mobile devices.
\newblock In \emph{Proceedings of the 2014 ACM SIGSAC Conference on Computer
  and Communications Security}, pages 1403--1414. ACM, 2014.

\end{thebibliography}

\end{document}